\documentclass[
twocolumn,
]{ceurart}

\AtBeginDocument{%
  }

\usepackage[inline]{enumitem}
\usepackage{multirow}
\usepackage{balance}
\usepackage{xspace}
\usepackage[capitalize,noabbrev]{cleveref}
\usepackage{amsmath,amsthm,amssymb}

\newtheorem{theorem}{Theorem}
\newtheorem{definition}[theorem]{Definition}

\newcommand{\pCTR}{$\mathsf{pCTR}$\xspace}
\newcommand{\pCVR}{$\mathsf{pCVR}$\xspace}
\newcommand{\pConvs}{$\mathsf{pConvs}$\xspace}

\begin{document}

\copyrightyear{2023}
\copyrightclause{Copyright for this paper by its authors.
  Use permitted under Creative Commons License Attribution 4.0
  International (CC BY 4.0).}

\conference{AdKDD'23}

\title{Private Ad Modeling with DP-SGD}

\author[2]{Carson Denison}
\author[1]{Badih Ghazi}[email=badihghazi@gmail.com]
\cormark[1]
\author[1]{Pritish Kamath}[email=pritishk@google.com]
\author[1]{Ravi Kumar}[email=ravi.k53@gmail.com]
\author[1]{Pasin Manurangsi}[email=pasin@google.com]
\author[1]{Krishna Giri Narra}[email=krishnanarra@google.com]
\author[1]{Amer Sinha}[email=amersinha@google.com]
\author[1]{Avinash V Varadarajan}[email=avaradar@google.com]
\author[1]{Chiyuan Zhang}[email=chiyuan@google.com]
\address[1]{Google}
\address[2]{Anthropic.  Work done while at Google.}

\begin{abstract}
A well-known algorithm in privacy-preserving ML is differentially private stochastic gradient descent (DP-SGD). While this algorithm has been evaluated on text and image data, it has not been previously applied to ads data, which are notorious for their high class imbalance and sparse gradient updates. In this work we apply DP-SGD to several ad modeling tasks including predicting click-through rates, conversion rates, and number of conversion events, and evaluate their privacy-utility trade-off on real-world ads datasets. Our work is the first to empirically demonstrate that DP-SGD can provide both privacy and utility for ad modeling tasks.
\end{abstract}


\maketitle

\section{Introduction}

With increasing focus on privacy on the Web and mobile apps, and given the signal loss due to cookie deprecation by several platforms, there has been a great need for privacy-preserving ML methods applied to ad prediction tasks. The most widely used predictive models in digital advertising are typically trained on user data pertaining to one or multiple sites/apps, and are used by ad technology providers (Ad Techs) to optimize the placement of digital ads.

Differential Privacy (DP)~\cite{DworkMNS06, DworkKMMN06} has emerged as a popular notion of privacy that is extensively studied in the research community and widely deployed in industrial applications, especially for training ML models with provable privacy guarantees. 
The main goal of DP training in ad modeling is to mitigate the privacy risks. For instance, the training examples for ad prediction models often depend on cross-site information. In the absence of privacy guardrails, the weights of the trained model could reveal, e.g., the browsing history of users. Such leakage of user information present in the training data is mitigated by DP training methods.
Intuitively, DP achieves a trade-off between privacy and utility by allowing statistical analysis and learning based on population-wide properties, while limiting the influence of (private) information from any individual user on the final output or model.

A training algorithm takes the set of examples as the input and produces the (trained weights of the) model as the output. For deep learning, various algorithms were proposed to privatize a learning pipeline, such as PATE~\citep{papernot2016semi,papernot2018scalable} and DP-FTRL~\citep{kairouz2021practical}. But the most widely used generic algorithm is DP stochastic gradient descent (DP-SGD)~\citep{ponomareva2023dp}, which goes back to the work of \citet{abadi2016deep}. We focus on DP-SGD in this paper. At a high level, DP-SGD works by clipping the norm of the per-example gradient to limit the influence of each example, and then adding Gaussian noise to the mini-batch averaged gradient to achieve DP. Since most deep neural network models are trained using SGD or variants such as Adam, DP-SGD can be adapted to any existing training pipeline with minimum modification by just replacing the optimizer. 

However, a direct application of DP-SGD can lead to significant utility (i.e., accuracy) loss and a large computational overhead, in practice. In fact, until recently, it was not clear if DP-SGD was suitable for large-scale deep learning. Recent studies and success stories mostly focused on  vision~\citep{kurakin2022toward,de2022unlocking} and text~\citep{anil2021large,ponomareva2021} problems. 

In this work, we present a systematic study of DP-SGD on ad prediction tasks. 
These tasks have highly unbalanced label distributions, categorical features with extremely sparse signals, and models with large embedding layers. Such properties make ads prediction more challenging than many other tasks.
These difficulties meant that DP-SGD was commonly considered infeasible except for trivially large privacy budgets.  In contrast, we demonstrate that it is possible to train private models with DP-SGD with only a small utility drop even in the high privacy regime. 
For example, for a click-through prediction task, the AUC loss is increased by only 15.8\% relative to a non-private baseline ($0.1943\rightarrow 0.2250$) even at a privacy budget of $\varepsilon=0.5$.
Furthermore, we show that with care, the computation and memory overheads of DP-SGD can be made almost identical to that of non-private training.  

We will discuss the ideas that make these results possible, such as large batch training, efficient per-example gradient norm bounding, and improved privacy accounting.  We also provide a comparison of DP-SGD to LabelDP~\citep{ghazi2021deep,malek2021antipodes}, a DP notion that protects only the labels and not the features. To the best of our knowledge, ours is the first systematic study on training large deep neural networks privately for ad prediction tasks. We hope our results serve as an optimistic example towards DP training of large ad prediction neural networks. We also hope our detailed studies provide useful information for practitioners to improve the utility and minimize the overhead of DP training.

\section{Background}
 
Let $\mathcal{A}$ be a (stochastic) training algorithm that produces a model given a labeled training set.  We call two training sets \emph{neighboring} if they differ on a single labeled example.  

\begin{definition}[Differential Privacy]
Let $\varepsilon \geq 0, \delta \in [0, 1]$.  A randomized training algorithm $\mathcal{A}$ is \emph{$(\varepsilon,\delta)$-differentially private (($\varepsilon, \delta$)-DP)} if for all $S\subseteq \operatorname{Range}(\mathcal{A})$ and all neighboring training sets $D, D'$, it holds that
\[
\operatorname{Pr}[ \mathcal{A}(D) \in S ] \leq e^{\varepsilon} \cdot \operatorname{Pr}[ \mathcal{A}(D') \in S ] + \delta.
\]
\end{definition}

DP-SGD~\citep{abadi2016deep} is the most widely used DP training algorithm for a deep learning pipeline. Let $f_\theta$ be a neural network with trainable weights $\theta$, and $\{(x_1,y_1),\ldots,(x_B,y_B)\}$ be a random mini-batch of training examples. Let $\ell_i=\ell(f_\theta(x_i),y_i)$ be the loss on the $i$th example and let $\bar{\ell}= \frac{1}{B}\sum_{i=1}^B\ell_i$ be the average loss.  Standard training algorithms compute the \emph{average gradient} $\nabla_\theta\bar{\ell}$ and update $\theta$ with an optimizer such as SGD or Adam.  In DP-SGD, the per-example gradients $\nabla_\theta\ell_i$ are computed, and then rescaled to have a maximum $\ell_2$-norm $C$. The average of the norm-bounded per-example gradients is perturbed by adding independent Gaussian noise to each coordinate, and is subsequently passed to the optimizer. The privacy parameters $\varepsilon$ and $\delta$ depend on the noise multiplier, and other parameters such as the batch size and training steps; they can be estimated via privacy accounting~\citep{abadi2016deep}.

\section{Summary of Main Results}

\begin{table}[]
    \centering
    \caption{DP-SGD results (average over five runs) on three different ads prediction datasets. Each row show the results under a specific privacy budget $\varepsilon$. It is common~\citep{abadi2016deep,li2021large,he2022exploring} to set $\delta=\mathcal{O}(1/N)$, where $N$ is the number of training examples. In this paper, we fix $\delta=1/N$. The percentages are relative loss increment calculated as $(L_\varepsilon - L_\infty)/L_\infty$, where $L_\epsilon$ is the loss for the $(\epsilon, \delta)$-DP model and $L_\infty$ is the loss for the non-private ($\varepsilon=\infty$) baseline. We use AUC loss (i.e., $1-\text{AUC}$) for \pCTR and \pCVR, and Poisson log loss for \pConvs.}
    \label{tab:main-results}
    \begin{tabular}{c|ccc}
    \toprule
      Privacy Budget & \multicolumn{3}{c}{Relative Loss Increment (\%)} \\
      \cline{2-4}
      ($\varepsilon$) & \pCTR & \pCVR & \pConvs \\
    \midrule
      0.5   & 16.11 & 9.99 & 97.04 \\
      1.0   & 13.58 & 9.51 & 85.71 \\
      3.0   &  8.77 & 8.55 & 68.19 \\
      5.0   &  7.40 & 7.84 & 67.14 \\
      10.0  &  6.27 & 7.28 & 60.64 \\
      30.0  &  5.67 & 6.45 & 46.00 \\
      50.0  &  5.56 & 5.84 & 41.20 \\
    \bottomrule
    \end{tabular}
\end{table}

We focus on three common predictions tasks for which Ad Techs build ML models.
\begin{itemize}[leftmargin=*]
    \item \pCTR: predict the click-through rate for an ad.
    \item \pCVR: predict the conversion rate for an ad click; here, only whether a conversion takes place matters, regardless of the number of conversions.
    \item \pConvs: predict the expected number of conversions after an ad click; this is a regression problem against integer count labels. 
\end{itemize}
We evaluate \pCTR on the public Criteo dataset~\citep{criteo}, and \pCVR, \pConvs on a proprietary dataset. We train the binary classification problems on \pCTR and \pCVR with the binary cross entropy loss and report the test \emph{AUC loss} (i.e., $1 - \text{AUC}$), and the regression problem on \pConvs with the Poisson log loss (PLL), and report the test PLL. Let $f_\theta(x)$ be the scalar prediction (i.e., the logit value) of the neural network, and $y$ be the integer counting label.  \emph{PLL} is defined as $\ell(f_\theta, (x,y)) := \exp(f_\theta(x)) - yf_\theta(x)$. In all our experiments with $(\varepsilon,\delta)$-DP, we set $\delta$ to be $1/N$, where $N$ is the number of training examples in the dataset.

The percentage increases in loss at various privacy budgets are presented relative to a non-private baseline\footnote{Our non-private AUC loss for the \pCTR task is 0.1943. We are unable to report the absolute non-private baseline losses for \pCVR and \pConvs on the proprietary datasets due to confidentiality.} in \Cref{tab:main-results}.
We found DP-SGD can properly train the models for all three tasks with a reasonable loss gap, even for very high privacy (e.g., $\varepsilon<1$) regimes. Furthermore, when implemented carefully, the computation and memory overheads can be minimized, allowing a training throughput similar to the non-private baseline.  Next, we present in detail the techniques that enabled optimal privacy-utility trade-off and minimum memory/computation overhead, respectively.

\section{Privacy-Utility Trade-off}
\label{sec:privacy-util-trade-off}

One of the main obstacles for adopting DP-SGD in real-world deep learning pipelines is the potential deterioration of the model performance.  Norm-rescaling  and Gaussian noise introduce, respectively, bias and variance to  gradient estimation. As a result, the trained models have lower performance. In this section, we study these challenges in detail by using the training setup from the non-private baseline as the starting point, and describe various improvements that lead to our main results. We state our results for Criteo \pCTR, but the key points also hold on other prediction tasks.

\begin{figure}
    \centering
    \includegraphics[width=\linewidth]{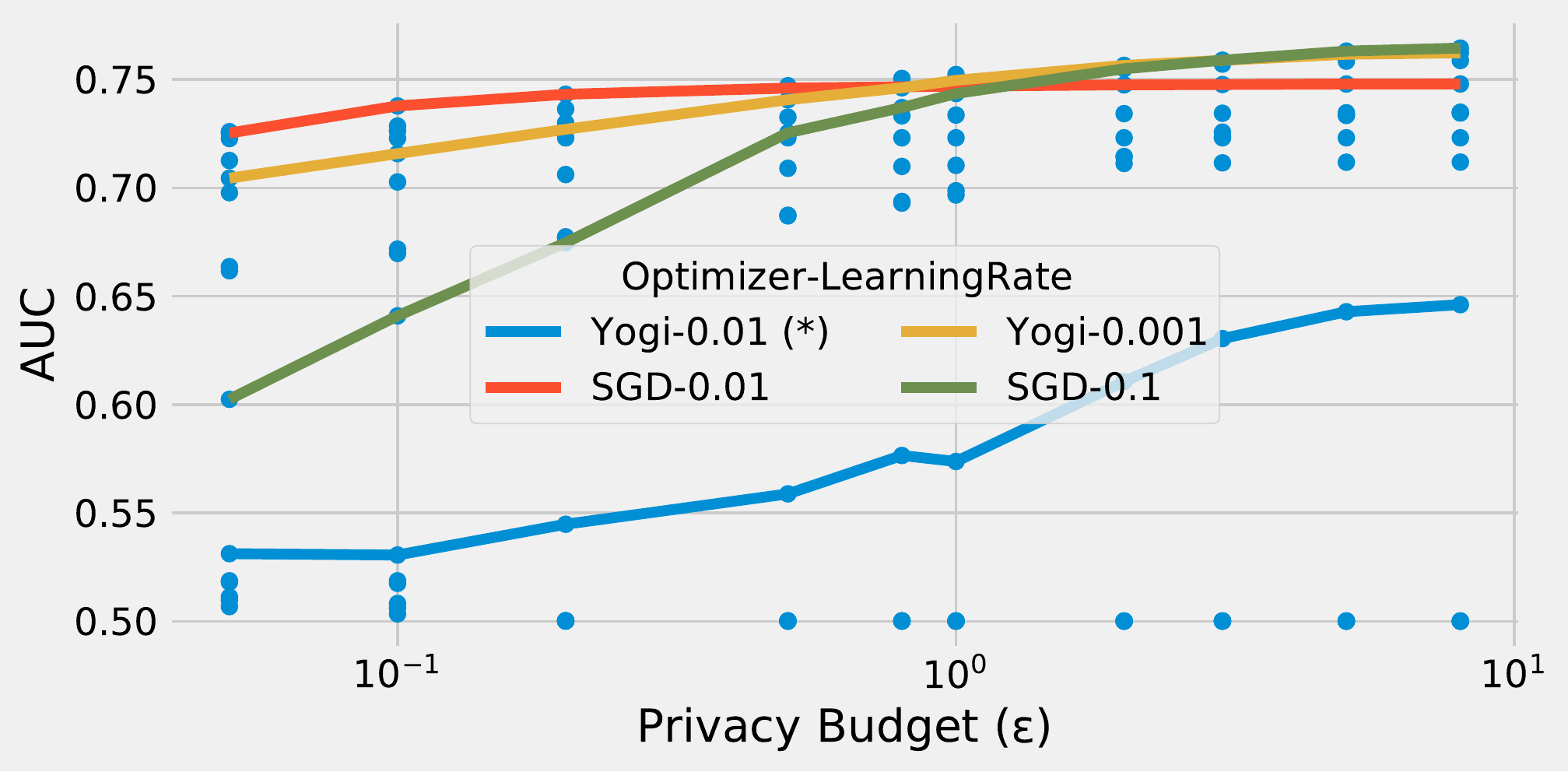}
    \caption{AUC under different (optimizer, learning rate) combinations on Criteo \pCTR. Each dot is the average test AUC of 5 random runs with a specific combination of $\varepsilon$, optimizer, and learning rate. We evaluate the following optimizers: SGD, Yogi~\citep{zaheer2018adaptive}, Adagrad~\citep{duchi2011adaptive}, Adam~\citep{kingma2014adam}, AdamW~\citep{loshchilov2017decoupled}; and learning rates: 0.001, 0.01, 0.1. The blue line (Yogi-0.01) is with the optimal hyperparameters used in the non-private baseline.}
    \label{fig:criteo-hp-tune}
\end{figure}

\subsection{Hyperparameter Tuning}

The hyperparameters for training the non-private baseline models in real-world applications are typically extensively tuned. However, those hyperparameters are not necessarily the best for training with DP.
\Cref{fig:criteo-hp-tune} plots the AUC of DP trained models under different optimizer and learning rate configurations. Specifically, the blue line shows that directly reusing the optimal hyperparameters for the non-private baseline leads to significant utility gap comparing to the best hyperparameters re-tuned under DP training. Note that the optimal hyperparameters also depend on $\varepsilon$: SGD with learning rate 0.1 performs better than with learning rate 0.01 in the low privacy (large $\varepsilon$) regime but worse in the high privacy (small $\varepsilon$) regime. For simplicity, we choose a single configuration (SGD-0.01) for the rest of the study; this already improves the AUC significantly compared to using the non-private hyperparameters\footnote{Ideally, hyperparameter tuning should be performed with DP, e.g., using~\citep{gupta2010differentially}. However, following most prior work on DP training, we ignore the DP cost for hyperparameter tuning.}. 

\subsection{Bias-Variance Trade-off}
\label{sec:bias-variance}

\begin{table}[]
    \caption{AUC under different per-example norm bound ($C$) and privacy budget  ($\varepsilon$) on Criteo \pCTR.}
    \label{tab:clip-norm}
    \centering
    \begin{tabular}{c|ccc}
\toprule
$\varepsilon$ & $C=1.0$ & $C=3.0$ & $C=30.0$\\
\midrule
$0.5$ & $.7498\pm .0011$ & $.7441\pm .0014$ & $.5000\pm .0000$\\
$3.0$ & $.7524\pm .0010$ & $.7610\pm .0010$ & $.6971\pm .0021$\\
$8.0$ & $.7528\pm .0010$ & $.7629\pm .0010$ & $.7473\pm .0009$\\
\bottomrule
    \end{tabular}
\end{table}

With fixed noise multiplier (and other hyperparameters like the training steps), the norm bound $C$ for each per-example gradient allows a bias-variance trade-off in the gradient estimation for a fixed $\varepsilon$. Specifically, increasing $C$ reduces bias but increases the variance in gradient estimation due to the addition of more noise.  A proper choice of $C$ improves the quality of trained model. For example, \Cref{tab:clip-norm} shows that in a low privacy regime (where the noise multiplier is smaller), increasing $C$ leads to better model performance because the added noise is already small. On the other hand, a large $C$ can hurt performance in the high privacy regime by requiring large amounts of noise. Adaptively choosing the norm bound $C$ has been studied in the literature~\citep{thakkar2019differentially,pichapati2019adaclip}.

\subsection{Large Batch Training}

Batch size is an important hyperparameter that affects different aspects of DP training. Specifically, increasing the batch size leads to a larger subsampling ratio, which implies larger noise multiplier for the same $\varepsilon$. On the other hand, increasing the batch size also reduces the effective noise that is added to the average gradient because the noised sum of gradients are divided by the batch size~\cite[Algorithm 1]{abadi2016deep}. Moreover, when fixing the number of training epochs, varying the batch size also changes the number of training steps, which affects both the privacy accounting and model utility.

\begin{figure}
    \centering
    \begin{tabular}{@{}c@{}c@{}}
    \includegraphics[width=.49\linewidth]{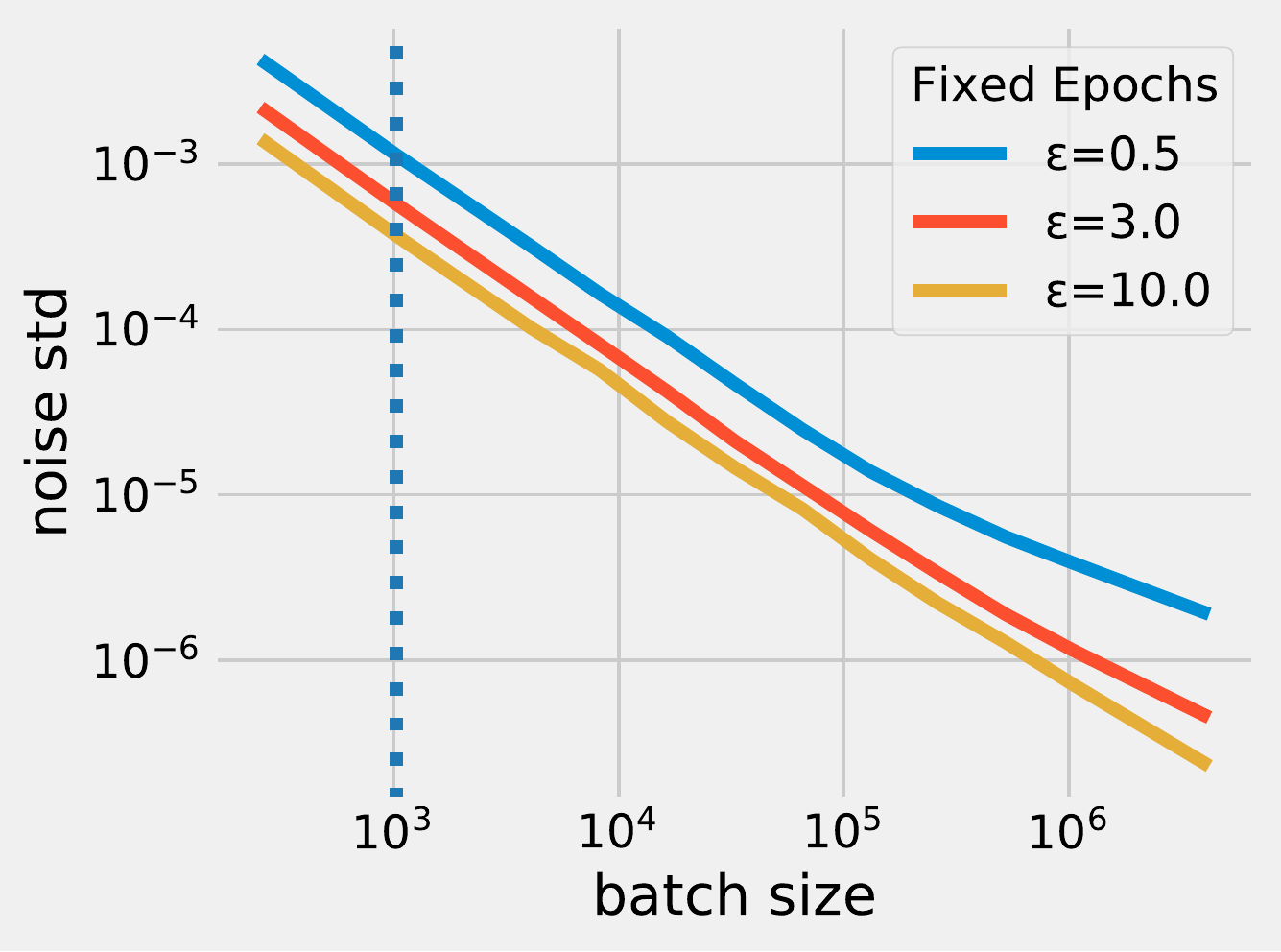} \; &
    \includegraphics[width=.49\linewidth]{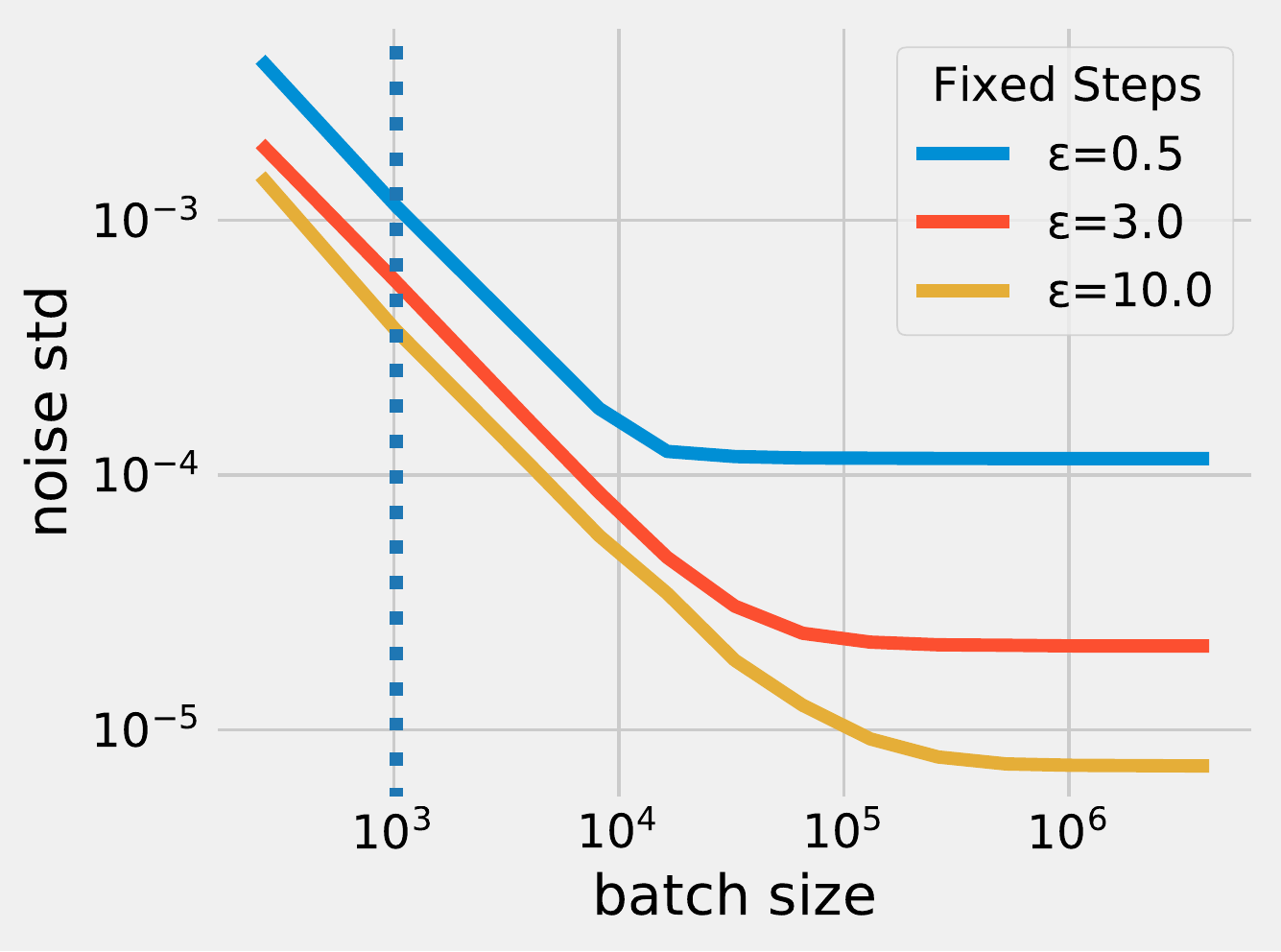} \\
    (a) Fixed epochs & (b) Fixed steps
    \end{tabular}
    \caption{Relationship between batch size and noise std for fixed \textbf{(a)}  training epochs and \textbf{(b)} training steps. The dotted line shows the batch size (1024) in the non-private baseline.}
    \label{fig:bs-std-relation}
\end{figure}

In \Cref{fig:bs-std-relation}, we plot the relationship between the  Gaussian noise standard deviation (std) required to guarantee a certain $\varepsilon$, when the batch size changes. In \Cref{fig:bs-std-relation}(a), we fix the total number of passes (epochs) over the data; the noise std continues to decrease as the batch size increases beyond $10^6$. However, this is not very realistic, because with fixed training epochs, increased batch sizes lead to decreased number of training steps. Modern neural networks trained with stochastic optimizers usually require a minimum number of steps to fit the data well. In \Cref{fig:bs-std-relation}(b), we plot the relation under the condition of fixed number of training steps. In this case, the benefit of large batch sizes plateaus.

\begin{table}
    \caption{AUC with large batch size and training epochs on Criteo \pCTR. The results can be compared with \Cref{tab:clip-norm} (batch size 1024, 8 epochs).}
    \label{tab:large-batch}
    \centering
    \begin{tabular}{crcc|c}
    \toprule
    $\varepsilon$ & batch size & epochs & $C$ & AUC \\
    \midrule
    $0.5$  & 16,384 & 64 & 30 & $0.7740\pm 0.0003$ \\
    $0.5$  & 8,192 & 32 & 30 & $0.7689\pm 0.0003$ \\
    \midrule
    $3.0$ & 16,384 & 64 & 30 & $0.7810\pm 0.0003$ \\
    $3.0$ & 8,192 & 32 & 30 & $0.7782\pm 0.0003$ \\
    \midrule
    $8.0$ & 16,384 & 64 & 30 & $0.7818\pm 0.0003$ \\
    $8.0$ & 8,192 & 32 & 30 & $0.7799\pm 0.0004$ \\
    \bottomrule
    \end{tabular}
\end{table}

Since SGD optimization is itself stochastic, reducing the noise std might not improve the model \emph{ad infinitum}. However, larger batches reduce noise, which allows the norm bound $C$ to be increased while keeping the total added noise and privacy level fixed.
Increasing $C$ can yield performance gains. Specifically, \Cref{tab:large-batch} shows the model performances on Criteo \pCTR with increased batch sizes, training epochs, and norm bound $C$. Comparing with \Cref{tab:clip-norm}, large batch sizes lead to significant performance boost. Note that for $\varepsilon=0.5$, the model cannot properly train ($\text{AUC}=0.5$) for $C=30$, but reaches $\text{AUC}=0.77$ when the batch size increases $16\times$.
We note that large batches have previously been used to improve DP-SGD for vision and language models~\citep{anil2021large, kurakin2022toward,de2022unlocking}. 

\subsection{Microbatching}

Microbatching was originally used to mitigate the computation and memory overhead of vanilla DP-SGD implementations~\citep{abadi2016deep}. It works by partitioning each mini-batch of $B$ training examples into ``microbatches'' of size $B_\mu$, and performing $\ell_2$-norm rescaling on the average gradient of each microbatch (as oppose to the per-example gradient). With large $B_\mu$, even a vanilla DP-SGD implementation could be quite efficient because microbatch average gradients can be computed with standard backpropagation API. However, because group norm bounding changes the sensitivity of the mean gradient query, the magnitude of Gaussian noises scale up by a factor of $B_\mu$ under the same privacy guarantee. As a result, large $B_\mu$ generally leads to worse model utility.

\begin{figure}
    \centering
    \begin{tabular}{@{}c@{}c@{}}
    \includegraphics[width=.375\linewidth]{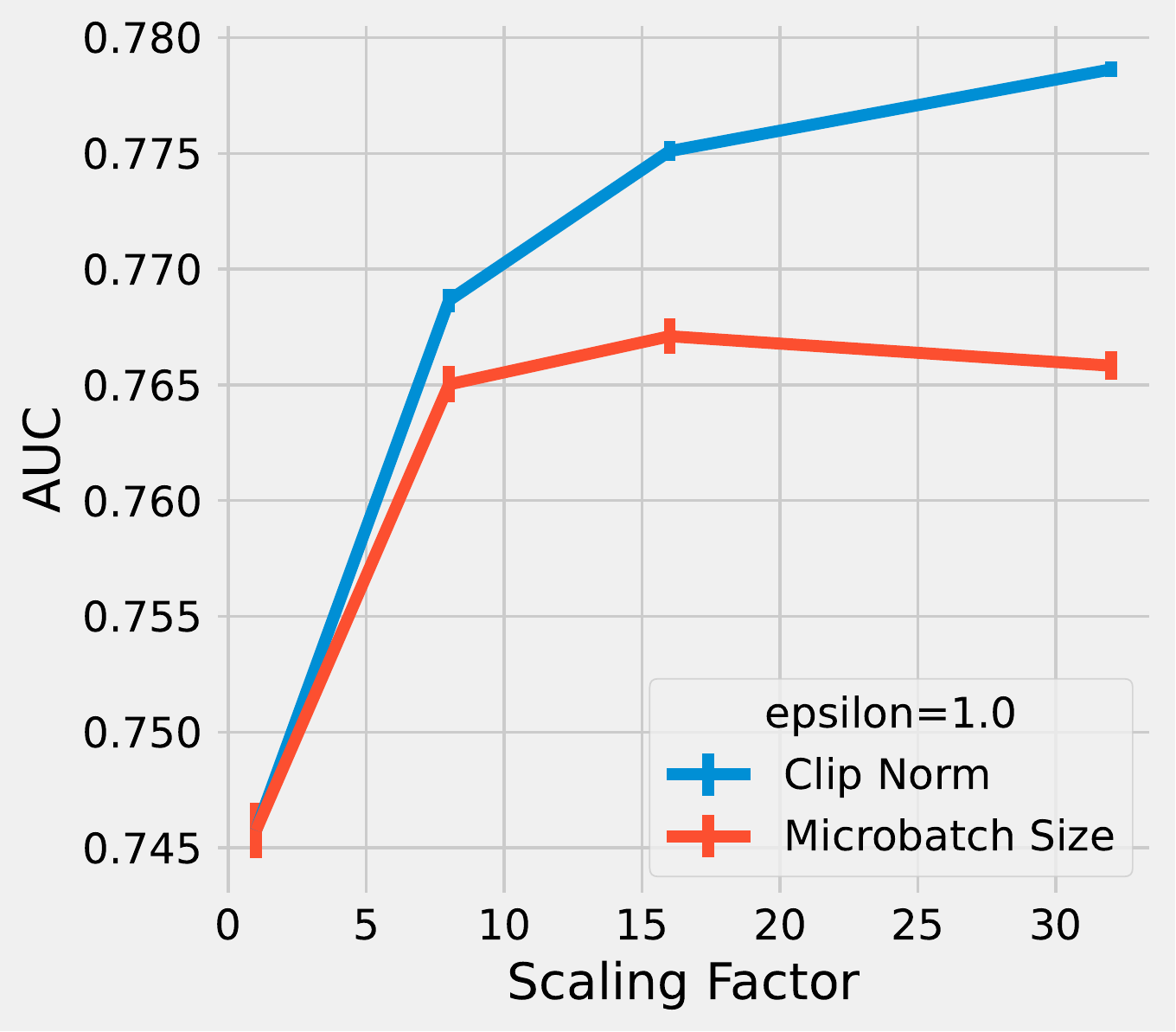} \; &
    \includegraphics[width=.61\linewidth]{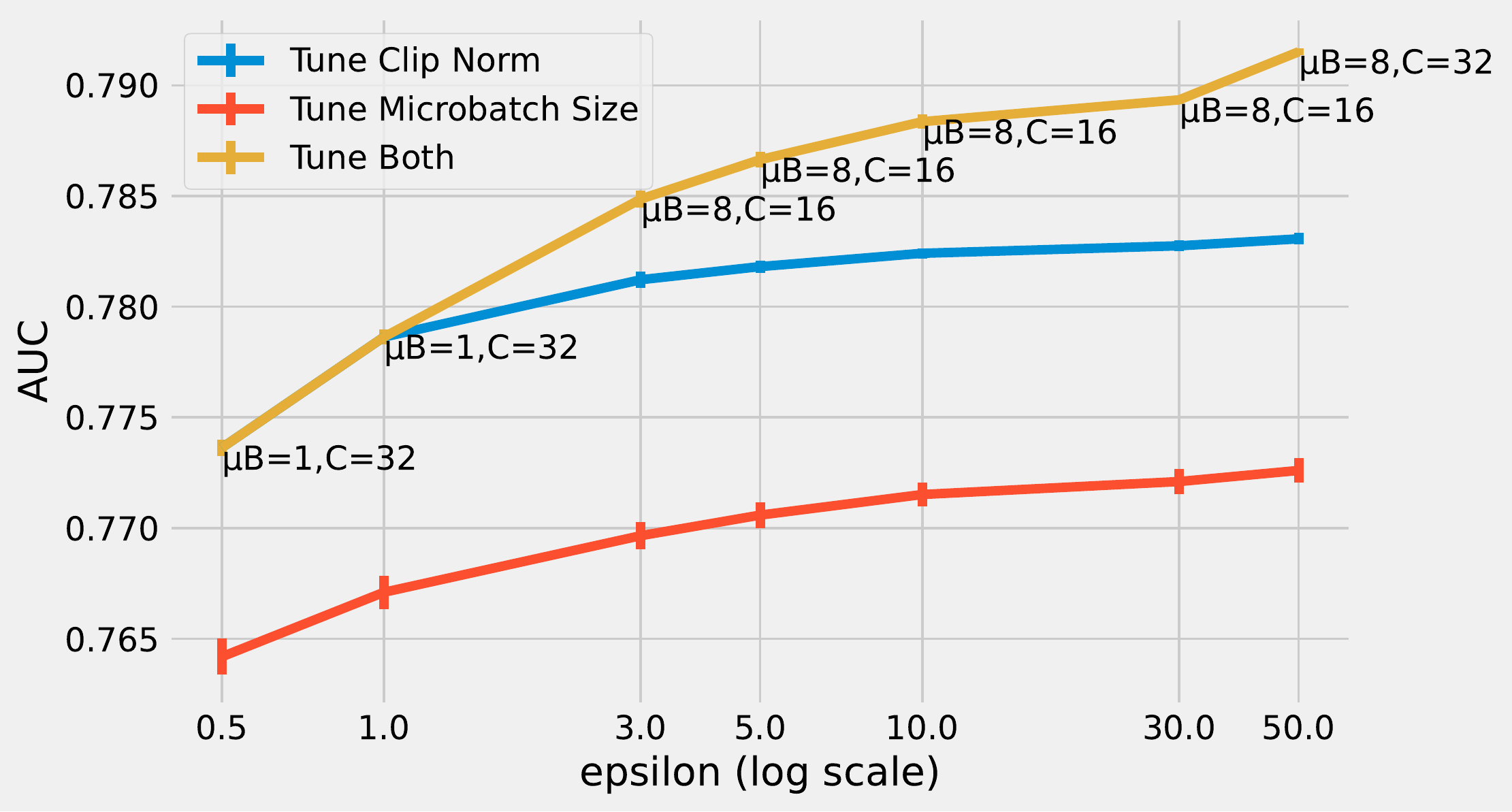} \\
    (a) $\varepsilon=1.0$ & (b) Tuning for each $\varepsilon$
    \end{tabular}
    \caption{The impact on utility by tuning the norm bound $C$ and microbatch size $B_\mu$ individually, and jointly, for Criteo \pCTR. The search range for both $C$ and $B_\mu$ are $\{1, 8, 16, 32\}$.}
    \label{fig:microbatching}
\end{figure}

In our case, with an efficient implementation (\Cref{sec:computation}), we do not need to use microbatching. However, we empirically found that small $B_\mu$ could improve the model utility. One potential reason is that this allows a different kind of bias-variance trade-off from changing the norm bound $C$ (\Cref{sec:bias-variance}). The increased noise leads to higher variance in gradient estimation. On the other hand, averaging the gradients within the microbatch before clipping could potentially reduce the bias when there are cancellations. For example, averaging $B_\mu$ i.i.d. Gaussian vectors reduces the expected norm by a factor of $1/\sqrt{B_\mu}$. The cancellation of gradient vectors is hard to characterize theoretically, but as shown in \Cref{fig:microbatching}(a), increasing $B_\mu$ moderately indeed improve the utility for Criteo \pCTR. Similar to $C$, the maximum tolerable value before it starts hurting the utility increases with $\varepsilon$. On the other hand, $C$ and $B_\mu$ seem to be helping with bias reduction in different ways, because increasing each of them by the same scaling factor leads to different AUC, even though the Gaussian noise scale (i.e., the variance) would be increased by the same amount. As a result, we could combine the two factors to further boost the utility, as verified in \Cref{fig:microbatching}(b).

\subsection{Tighter Privacy Accounting}

Privacy accounting estimates the privacy budget $\varepsilon$ for a DP-SGD trained model according to the specific hyperparameters such as the noise std, the norm bound $C$, the number of training steps, etc. R\'enyi Differential Privacy (RDP) accounting has been the most widely used approach in DP-SGD since the original paper~\cite{abadi2016deep}. With RDP, $\varepsilon$ can be computed using only the number of epochs, the batch size, and the noise std.

In this paper, we explore latest advances in privacy accounting to provide tighter estimates. Specifically, several recent works \cite{meiser2018tight, sommer2019privacy} studied numerical methods for estimating the privacy parameters of a DP mechanism to an arbitrary accuracy using the notion of \emph{privacy loss distributions} (PLD).  A crucial property is that the PLD of a composition of multiple mechanisms is the convolution of their individual PLDs. Thus, \citet{koskela2020computing} used the Fast Fourier Transform (FFT) in order to speed up the computation of the PLD of the composition; faster algorithms and more accurate algorithms have been proposed in subsequent work \cite{gopi21numerical,ghazi22faster}, and have been the basis of multiple open-source implementations from both industry and academia including \cite{DPBayes, GoogleDP, MicrosoftDP}. In particular, in this paper, we use the so-called ``connect-the-dots'' algorithm of \citet{doroshenko22connect} for privacy accounting.

\begin{figure}[t]
     \centering
     \begin{tabular}{@{}c@{}c@{}}
     \includegraphics[width=.51\linewidth]{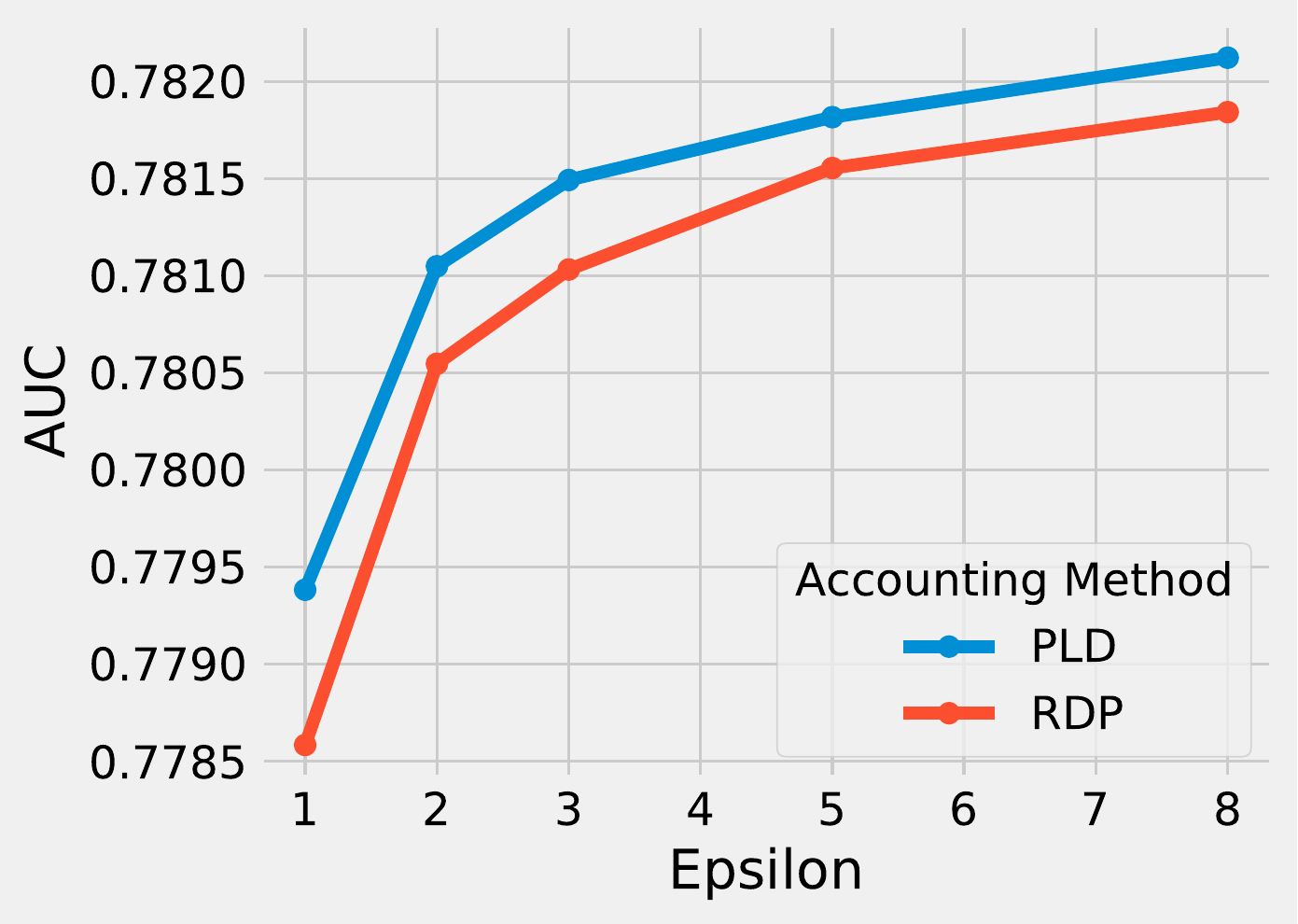} \; &
     \includegraphics[width=.48\linewidth]{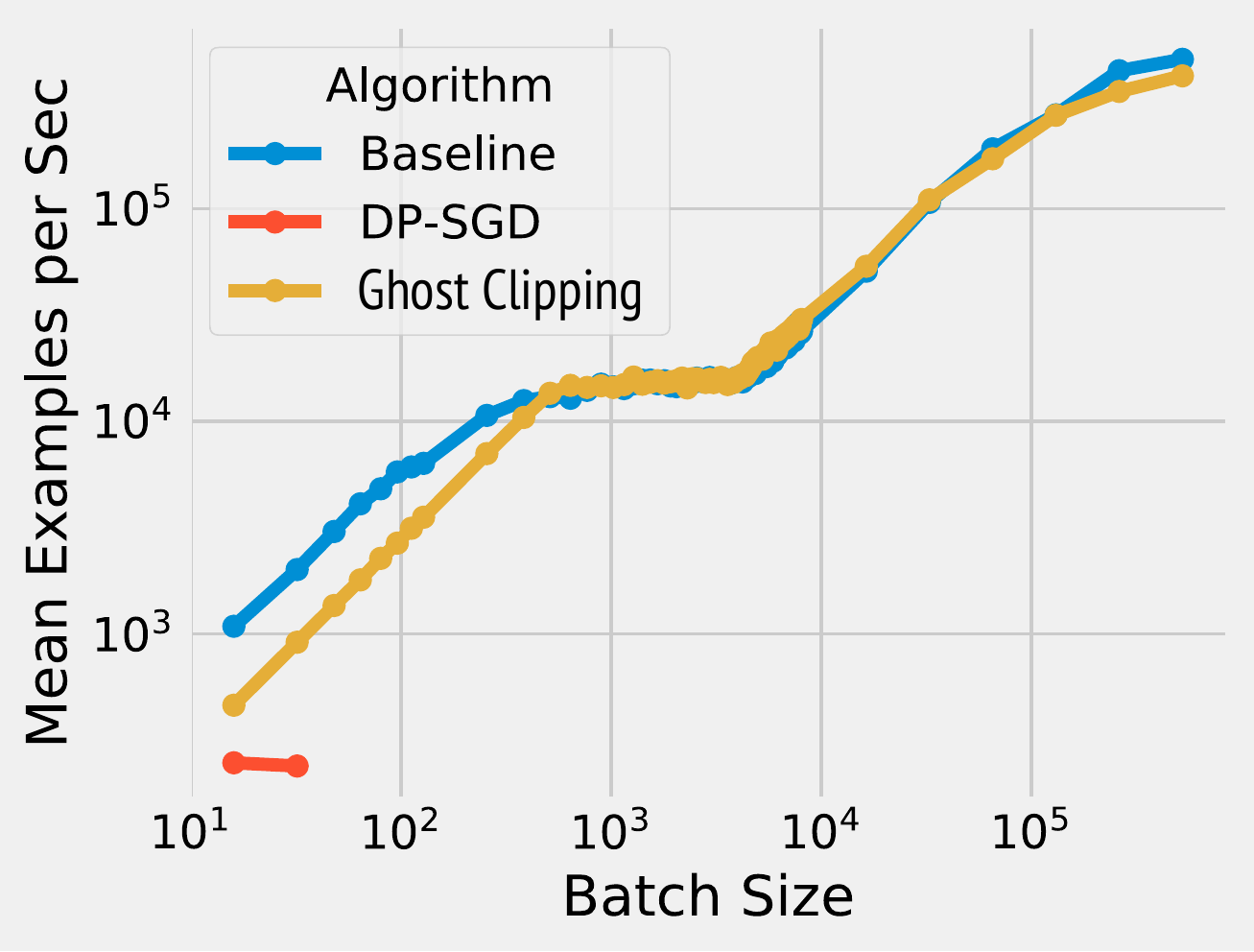} \\
     (a) Accounting & (b) Computation
     \end{tabular}
    \caption{\textbf{(a)} Comparison between RDP and PLD accounting and \textbf{(b)} comparison of computational efficiency. Measured on Criteo \pCTR with identical model architecture, hyperparameters one a single Nvidia\textsuperscript{\textregistered} Tesla\textsuperscript{\textregistered} P100 GPU.}
    \label{fig:pld-speed-combo}
\end{figure}

\Cref{fig:pld-speed-combo}(a) compares RDP accounting to the improved PLD accounting. Because the PLD estimation is tighter, a given $\varepsilon$ requires smaller noise than implied by  RDP accounting. We observe consistent improvements of model utility across all privacy regimes, with a larger gap for smaller $\varepsilon$'s.

\section{Computation \& Memory Overhead}
\label{sec:computation}

Another obstacle to using DP-SGD in real-world deep learning systems is the potential computation and memory overhead.  A naive implementation of DP-SGD that explicitly computes each per-example gradient can lead to several orders of magnitude more memory consumption and computational cost. Therefore, efficient implementations of DP-SGD have been studied from various angles, including micro-batching~\citep{abadi2016deep}, layer-specific algorithms~\citep{goodfellow2015efficient,lee2021scaling}, just-in-time compilation~\citep{subramani2021enabling}, and approximation via random projections~\citep{bu2021fast}.

\begin{table}[t]
    \caption{Comparison of memory consumption via the maximum batch size that can be trained on Criteo \pCTR with one Nvidia\textsuperscript{\textregistered} Tesla\textsuperscript{\textregistered} P100 GPU (`-' indicates that the training process ran out of memory.) }
    \label{tab:comp-and-mem}
    \centering
    \begin{tabular}{c|ccc}
    \toprule
    \multirow{2}{*}{Batch Size} & \multicolumn{3}{c}{Number of steps per second} \\
    \cline{2-4}\rule{0pt}{10pt}
    & Baseline & DP-SGD & Ghost Clipping\\
    \midrule
    32 & $62.73 \pm .08$ & $7.48 \pm .08$ & $28.57 \pm .61$\\
    64 & $63.91 \pm .71$ & - & $28.05 \pm .33$\\
    256 & $41.60 \pm .70$ & - & $27.52 \pm .35$\\
    1,024 & $14.17 \pm .83$ & - & $14.07 \pm .09$\\
    4,096 & $3.77 \pm .16$ & - & $3.88 \pm .26$\\
    16,384 & $3.10 \pm .34$ & - & $3.27 \pm .28$\\
    65,536 & $2.91 \pm .13$ & - & $2.62 \pm .24$\\
    524,288 & $0.96 \pm .05$ & - & $0.80 \pm .06$\\
    1,048,576 & - & - & - \\
    \bottomrule
    \end{tabular}
\end{table}

Here we demonstrate that when implemented with care, DP-SGD can be run with small computation and memory overheads for ads prediction models. This is enabled by the following observations: 
\begin{enumerate*}
    \item To bound sensitivity, we only need per-example gradient norms, \emph{not} per-example gradient vectors.
    \item Once per-example gradient norms are computed, the norm-bounded average gradients can be computed using standard backpropagation with reweighted loss.
    \item Most ads prediction models only use embedding and linear layers. As first noted by \citet{goodfellow2015efficient}, the per-example gradient norms for fully connected layers can be efficiently estimated with standard backpropagation \emph{sans} materializing the per-example gradient vectors. Since an embedding layer is equivalent to a fully connected layer on one-hot encoded inputs, Goodfellow's observation can be used in DP-SGD.  
\end{enumerate*}

Based on these observations, we implement a two-pass algorithm that computes
per-example gradient norms in the first pass and the norm-bounded average gradient in the second pass. Since most of the computation and memory overhead comes from materializing per-example gradient vectors, our implementation is able to reduce these significantly. This technique is usually called ``ghost clipping'', and is also effective in private training for computer vision~\citep{bu2022scalable} and natural language processing~\citep{li2021large,he2022exploring}. We implement this algorithm in JAX~\citep{jax2018github}, and compare to a baseline implementation using JAX just-in-time (JIT) compilation, which is already faster than a vanilla implementation as reported in \citet{subramani2021enabling}, as well as the non-private baseline (called Baseline).

\Cref{fig:pld-speed-combo}(b) plots the number of examples per second using different training algorithms and batch sizes. We observe that Fast-DP-SGD scales similarly to the non-private baseline, and the computation overhead is negligible for intermediate to large batch sizes.  On the other hand, the JIT based implementation is not only slower but also incapable of handling batch sizes larger than 64 because its memory overhead grows linearly with the batch size. The maximum batch size that each algorithm can handle on a single GPU is documented in \Cref{tab:comp-and-mem}. We can see that the batch size cap is the same for Fast-DP-SGD and the non-private baseline, demonstrating that the memory overhead is negligible.

The experiments above were done on a single GPU.  The batch size can be easily scaled beyond the maximum value in \Cref{tab:comp-and-mem} by using multi-device data parallelism, and gradient accumulation across multiple backpropagation steps.

\section{Comparison to Label DP}

\begin{figure}[t]
    \centering    \includegraphics[width=\linewidth]{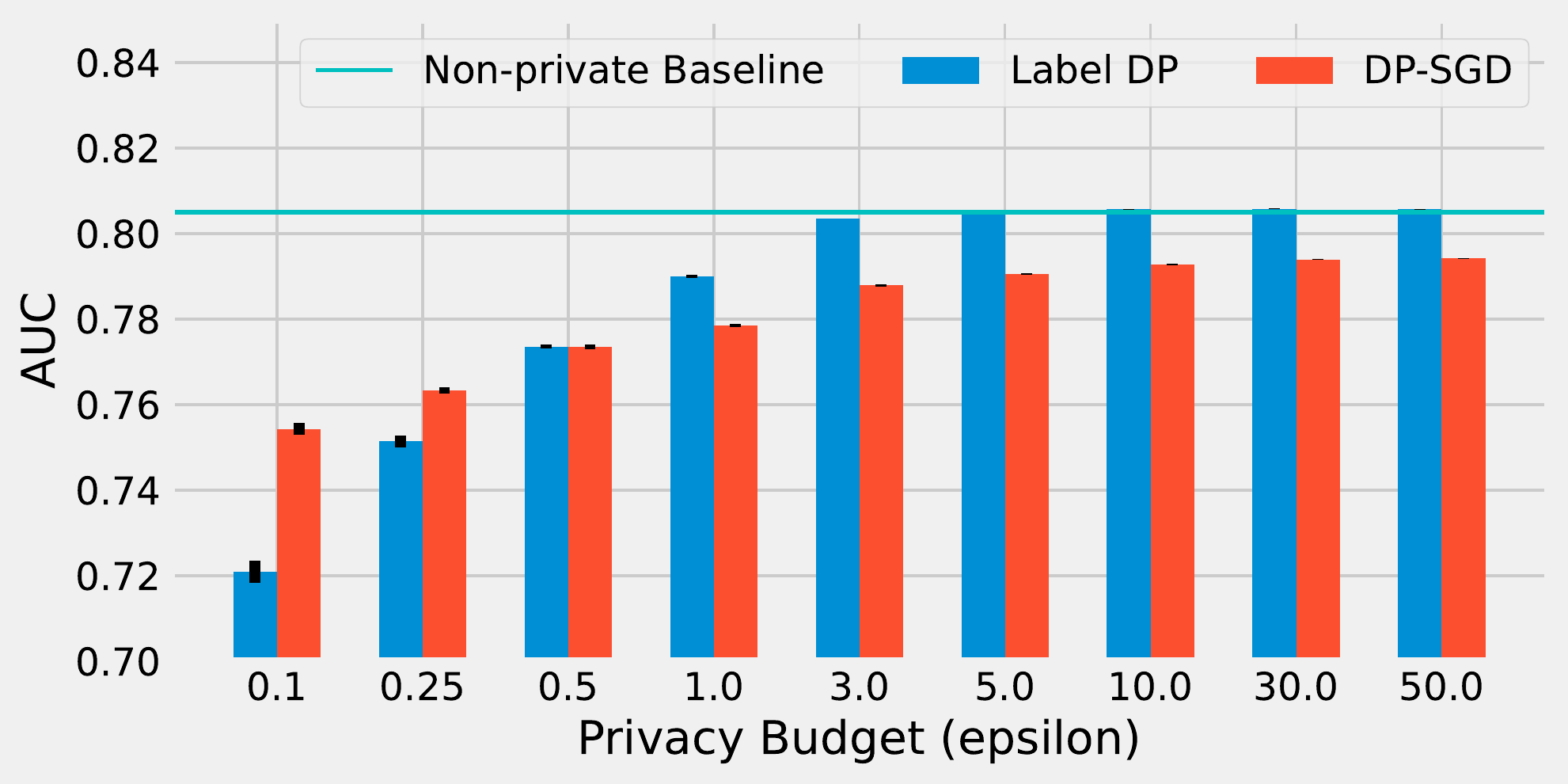}
    \caption{Comparing LabelDP (Randomized Response) with DP-SGD on Criteo under the same privacy budget $\varepsilon$.}
    \label{fig:dpsgd-vs-labeldp}
\end{figure}

Label differential privacy (LabelDP)~\citep{chaudhuri2011sample} is a notion where the features are public and only the labels need privacy protection. 
 It has been recently studied in deep learning~\citep{ghazi2021deep,malek2021antipodes}, and more specifically in ad modeling~\citep{ghazi2022regression}. In this section, we compare DP-SGD with LabelDP algorithms under the same privacy budget $\varepsilon$.  (Note that this is not an apples-to-apples comparison since unlike DP-SGD, LabelDP  protects only the labels; furthermore, since we use randomized response as our LabelDP algorithm, we have $\delta=0$.)  

From \Cref{fig:dpsgd-vs-labeldp}, we observe that LabelDP generally provides higher utility in low privacy (large $\varepsilon$) regimes, while DP-SGD outperforms it in high privacy (small $\varepsilon$) regimes. The behavior in high privacy regimes is counter-intuitive because DP-SGD has stronger privacy guarantees yet provides better utility.

\section{Conclusions}

In this work we showed that it is possible to privately train ad models using DP-SGD, while neither significantly sacrificing utility nor incurring computational cost.  An interesting research direction is to develop new private training algorithms for ``hybrid DP''---an interpolation between vanilla DP and Label DP in which some but not all of the features are public.  Another line of work would be to study if the low-rank nature / sparsity of the gradients can be exploited to reduce the noise needed for private training.

\subsection*{Acknowledgments}
We thank Silvano Bonacina and Samuel Ieong for many useful discussions and Christina Ilvento and Andrew Tomkins for their comments.

\bibliography{refs}

\appendix

\section{Training Details}

The Criteo \pCTR dataset~\citep{criteo} contains around 40 million examples. The raw dataset comes with a labeled training set and an unlabeled test set. We split the raw training set chronologically into a 80/10/10 partition of train/validation/test sets. The reported metrics are on this labeled test split. Each example consists of 13 integer features, 26 categorical features, and one binary label. We preprocess each integer feature with a log transformation.

The neural network consists of six layers. In the first layer, each categorical feature is mapped into a dense feature vector via an embedding layer. The embedding dimension is decided via a heuristic rule as $\mathsf{int}[ 2 V^{0.25}]$, where $V$ is the number of unique tokens in each categorical feature. The dense features are then concatenated with the log-transformed integer features to form the first layer representation. This representation are fed into four fully connected layers, each with an output dimension of $598$ and a ReLU activation function. Finally, a fully connected layer is used to compute the scalar prediction (the logit). There are 
around 78M trainable parameters in this neural network model.

The network is trained with binary cross entropy loss. Unless otherwise specified, we train the network for five  epochs, and we scale the base learning rate with a cosine decay~\citep{loshchilov2016sgdr}. In the non-private baseline, we use the Yogi optimizer~\citep{zaheer2018adaptive} with a base learning rate of 0.01, and a batch size of 1024.

Directly reusing the same hyperparameters for private training leads to suboptimal results. We discuss how to achieve better privacy-utility trade-off by adjusting different hyperparameters in \Cref{sec:privacy-util-trade-off}. Even though the analysis shows that optimal hyperparameters could be different for different range of privacy budget $\varepsilon$'s, for simplicity, we use a fixed value for most of hyperparameters in the final results of Criteo \pCTR in \Cref{tab:main-results}: SGD optimizer with base learning rate 0.01, momentum 0.9, batch size 65,536, and training for 150 epochs. We only tune the norm bound $C\in\{1.0, 10.0, 50.0, 100.0\}$ and microbatch size $B_\mu\in\{1,4,8\}$ for each $\varepsilon$ separately. 

The model and training setup for the \pCVR and \pConvs are similar, except that the \pConvs come with integer labels, thus are trained and evaluated with a Poisson log loss (PLL). Specifically, let $f_\theta(x)$ be the scalar prediction from the final layer of the neural network, interpreting $\exp(f_\theta(x))$ as predicting the mean of a Poisson distribution, then the log-likelihood of the integer label $y$ is
\[
yf_\theta(x) - \exp(f_\theta(x)) - \log (y!).
\]
The maximum (log-)likelihood training objective is thus equivalent to minimizing the following PLL function:
\begin{equation}
\ell(f_\theta, (x,y)) = \exp(f_\theta(x)) - yf_\theta(x).
\end{equation}
Note the $\log(y!)$ term is dropped since it is independent of the trainable parameters $\theta$. Therefore this is an \emph{unnormalized} PLL.

\end{document}